\documentclass{article}

\usepackage{amsfonts}
\usepackage{amsmath}
\usepackage{parskip}
\usepackage{cancel}
\usepackage{algorithm}
\usepackage{booktabs}
\usepackage{hyperref}
\usepackage[noend]{algorithmic}
\usepackage{bm}
\usepackage{graphicx}
\usepackage{float}
\usepackage{subfigure}

\hypersetup{
	colorlinks=true,
	allcolors=black,
}

\newcommand{\Cov}{\mathbf{Cov}}
\newcommand{\Var}{\textnormal{Var}}
\newcommand{\tr}{\textnormal{tr}}
\newcommand{\sgn}{\textnormal{sgn}}
\newcommand{\norm}[1]{\left\lVert #1 \right\rVert}

\title{An Isometric Stochastic Optimizer}
\author{Jacob Jackson \\ jacob@jacobjackson.com}

\date{July 24, 2023}

\begin{document}

\maketitle
\begin{abstract}
The Adam optimizer is the standard choice in deep learning applications. I propose a simple explanation of Adam's success: it makes each parameter's step size independent of the norms of the other parameters. Based on this principle I derive Iso, a new optimizer which makes the norm of a parameter's update invariant to the application of any linear transformation to its inputs and outputs. I develop a variant of Iso called IsoAdam that allows optimal hyperparameters to be transferred from Adam, and demonstrate that IsoAdam obtains a speedup over Adam when training a small Transformer.
\end{abstract}

\section{Introduction}

Recent years have seen unprecedented advances in machine understanding of language, images, audio, and video, driven in large part by the development of the Transformer in 2018 \cite{vaswani2017attention}. Yet despite this recent change in the dominant architecture, despite the growing economic incentive to speed up the training of large models, and despite dozens of papers claiming to have achieved such speedups through new optimization techniques, the most popular optimizer in practice remains Adam \cite{kingma2017adam}, which was published in 2015 and is itself closely related to RMSProp \cite{Tieleman2012} from 2012.


In this paper, I propose a simple explanation of Adam's success: it makes each parameter's step size independent of the norms of the other parameters. Based on this principle I derive Iso, a new optimizer.

In current large models, weight matrices constitute over 99\% of parameters \cite{brown2020language}. Therefore, I frame the stochastic optimization problem as follows: there is a linear layer defined by $W \in \mathbb{R}^{n \times m}$. The layer receives a vector $x \in \mathbb{R}^n$ as input, which is a random vector with an unknown distribution. The layer's output is computed as $y = W^\top x$. It receives a gradient $g \in \mathbb{R}^m$, also a random vector with an unknown distribution, representing the gradient of the loss with respect to $y$. The problem facing a stochastic optimizer is to choose an update $W_U$ for the matrix $W$ such that assigning $W \leftarrow W - \alpha W_U$ will decrease the loss.

The simplest stochastic optimizer is stochastic gradient descent (SGD), which uses the following update:
\begin{equation}
W_U = \mathbb{E}[xg^\top]
\end{equation}

This paper proposes Iso (short for \textit{isometric stochastic optimizer}), which uses the following update instead:
\begin{align}
W_U &= \mathbb{E}[xx^\top]^{-1/2} \, \mathbb{E}[xg^\top] \, \mathbb{E}[gg^\top]^{-1/2} \\
&= \Cov(x)^{-1/2} \, \mathbb{E}[xg^\top] \, \Cov(g)^{-1/2}
\end{align}

The Iso update can be seen as a multivariate extension of the formula for the correlation coefficient between two scalar random variables:
\begin{equation}
	\rho_{x,y} = \frac{\mathbb{E}[xy]}{\sqrt{\Var(x)\Var(y)}}
\end{equation}

In Section \ref{derivation}, I derive Iso as the update rule which makes the Frobenius norm of the update invariant to the application of any linear transformation to $x$ or $g$. In Section \ref{analysis}, I analyze Iso in comparison to Adam. In Section \ref{experiments}, I present evidence suggesting Iso works in practice. In Section \ref{scalability}, I consider the scalability of Iso to large models.

\begin{algorithm}
\begin{algorithmic}\itemindent=-10pt
\STATE Weight matrix $W \in \mathbb{R}^{n \times m}$
\STATE Learning rate $0 < \alpha$
\STATE Momentum decay $0 \le \beta < 1$
\STATE Initialize momentum $M \in \mathbb{R}^{n \times m} = 0$
\STATE Initialize left covariance $L \in \mathbb{R}^{n \times n} = 0$
\STATE Initialize right covariance $R \in \mathbb{R}^{m \times m} = 0$
\FOR{each iteration}
  \STATE Receive inputs $X \in \mathbb{R}^{b \times n}$ and output gradients $G \in \mathbb{R}^{b \times m}$
  \STATE $M \leftarrow M + (1 - \beta)(X^\top G - M)$
  \STATE $L \leftarrow L + (1 - \beta)(X^\top X - L)$
  \STATE $R \leftarrow R + (1 - \beta)(G^\top G - R)$
  \STATE $W \leftarrow W - \alpha L^{-1/2} M R^{-1/2}$
\ENDFOR
\end{algorithmic}
\caption{Iso with momentum.}
\end{algorithm}

\section{Derivation}
\label{derivation}

\subsection{The pure noise problem}
Consider the following problem, which I call the \textit{pure noise problem}: we have a linear model defined by a sequence of $n \times n$ weight matrices $W_1,\dots,W_k$. We receive input $x \sim \mathcal{N}(0, I_n)$ and compute the model output as:
\begin{equation}
	y = (W_1 W_2 \dots W_k)^\top x
\end{equation}
The loss function is $L(y) = y^\top z$, where $z \sim \mathcal{N}(0, I_n)$, so that the gradient with respect to $y$ is $z$. Note that this is a different problem than regression with $x$ as input and $z$ as output: in that case the gradient would be $y - z$ rather than $z$. This is why I call it the ``pure noise problem'': the gradient is pure noise with no relationship to $x$.

What can be expected of a stochastic optimizer on such a problem? Certainly it can't be expected to learn anything, since the gradient is noise. However, we could reasonably expect it to keep the model parameters within a bounded range, without sending them to infinity or zero. But this is not what SGD does. When run on the pure noise problem with 2 more layers, SGD sends the weight norm to infinity at an exponential rate, regardless of the learning rate or initialization.\footnote{Weight decay doesn't help: with sufficiently strong weight decay, the model avoids diverging, but at the cost of sending the weights to 0 instead. The weight norm is still exponential in the timestep, just with base less than 1.}

The problem is caused by a positive feedback loop. Consider the case $k=2$, where $y = W_2^\top W_1^\top x$. The update to $W_1$ is $x z^\top W_2^\top$, and the update to $W_2$ is $W_1^\top x z^\top$. Since there is no source of negative feedback, the variance of the weights grows with the timestep $t$. By itself this is not a problem (the same is true when $k=1$, but there is no exponential growth in that case). The problem is that increase in the norm of $W_1$ causes the updates to $W_2$ to become larger, and vice versa. This positive feedback causes exponential growth of the weight norm.

In practical problems, there is negative feedback arising from the dependence of the gradient on the model output, which prevents divergence. However, the learning rate must be set low enough for this negative feedback to take effect, which limits the step size that can be used with SGD.

In general, the dependence of each weight matrix's step size on the norms of other parameters makes models hard to tune because parameter norms typically change during training: a parameter may have small norm at initialization and large norm once the model reaches its steady state, or vice versa. With SGD, this change in norm will implicitly increase or decrease the step sizes of other parameters, which is not desirable.

Therefore, I propose the following principle for optimizing neural networks, which I call the ``invariant step size principle'':
\begin{center}
\textit{The step size for a weight matrix should be invariant to linear transformation of its inputs and outputs.}
\end{center}
This is a simple way of breaking the relationship between each parameter's step size and the norms of the other parameters. Note that Adam follows this principle, but only for diagonal transformation of inputs and outputs.

\subsection{Derivation of Iso}
I formalize the principle described above as follows. Let $W \in \mathbb{R}^{n \times m}$ be a weight matrix. Let $X \in \mathbb{R}^{b \times n}$ be the matrix of inputs to $W$, where $b$ is the batch dimension, and let $G \in \mathbb{R}^{b \times m}$ be the matrix containing the gradient of the loss with respect to $XW$, so that the SGD update would be $X^\top G$. We want to choose an update $W_U(X, G)$ such that if $A \in \mathbb{R}^{n \times n}$ and $B \in \mathbb{R}^{m \times m}$ are invertible matrices, then $\norm{W_U(X, G)}_F = \norm{W_U(XA, GB)}_F$.

We assume the solution has the form of multiplication by preconditioning matrices $L_X$ and $R_G$: $W_U(X, G) = L_X^\top X^\top G R_G$.

We proceed using the matrix identities $\norm{A}_F^2 = \tr(A A^\top)$ and $\tr(AB) = \tr(BA)$:
\begin{align}
\norm{W_U(XA, GB)}_F^2 &= \norm{L_{XA}^\top A^\top X^\top G B R_{GB}}_F^2 \\
&= \tr(L_{XA}^\top A^\top X^\top G B R_{GB} R_{GB}^\top B^\top G^\top X A L_{XA}) \\
&= \tr(G B R_{GB} R_{GB}^\top B^\top G^\top X A L_{XA} L_{XA}^\top A^\top X^\top)
\end{align}
The choice of $L_{XA}$ and $R_{GB}$ that allows us to cancel $A$ and $B$ is $L_{XA} = (A^\top X^\top X A)^{-1/2} = \Cov(XA)^{-1/2}$ and $R_{GB} = (B^\top G^\top G B)^{-1/2} = \Cov(GB)^{-1/2}$.
\begin{align}
& \,\,\,\,\,\,\,\, \tr(G B R_{GB} R_{GB}^\top B^\top G^\top X A L_{XA} L_{XA}^\top A^\top X^\top) \\
&= \tr(G B (B^\top G^\top G B)^{-1} B^\top G^\top X A (A^\top X^\top X A)^{-1} A^\top X^\top) \\
&= \tr(G \cancel{B B^{-1}} (G^\top G)^{-1} \cancel{(B^\top)^{-1} B^\top} G^\top X \cancel{A A^{-1}} (X^\top X)^{-1} \cancel{(A^\top)^{-1} A^\top} X^\top) \\
&= \tr(G (G^\top G)^{-1} G^\top X (X^\top X)^{-1} X^\top) \\
&= \tr(G R_G R_G^\top G^\top X L_X L_X^\top X^\top) \\
&= \norm{L_X^\top X^\top G R_G}_F^2 \\
&= \norm{W_U(X, G)}_F^2
\end{align}
This shows that the norm of the Iso update is invariant to linear transformation of its inputs and outputs.

Note that the update can change when the inputs are transformed: only the norm is invariant. For example, Iso is equivariant to orthogonal transformations: if $Q$ and $U$ are orthogonal, then $W_U(XQ, GU) = Q^\top W_U(X, G) U$.

\section{Analysis}
\label{analysis}

In this section, I give theoretical reasons to prefer Iso to Adam \cite{kingma2017adam}, which is currently the most popular optimizer.

When optimizing a weight matrix $W \in \mathbb{R}^{n \times m}$ with inputs $X$ and output gradients $G$ as defined in the previous section, Adam maintains a matrix $M \in \mathbb{R}^{n \times m}$ of first moments and a matrix $V \in \mathbb{R}^{n \times m}$ of second moments, which are updated according to the following rules (I omit the bias correction term for simplicity):
\begin{align}
M &\leftarrow M + (1 - \beta_1)(X^\top G - M) \\
V_{ij} &\leftarrow V_{ij} + (1 - \beta_2)((X^\top G)_{ij}^2 - V_{ij}) \\
(W_U)_{ij} &= \frac{M_{ij}}{\sqrt{V_{ij}} + \varepsilon} \label{adamupdate}
\end{align}

\subsection{Sign descent}
There is a well-known \cite{bernstein2018signsgd,kunstner2023noise,balles2020dissecting} connection between Adam and sign descent, a stochastic optimizer which uses the following update rule:
\begin{equation}
	W_U = \sgn(X^\top G)
\end{equation}
In fact, Adam is equivalent to sign descent in the limit as batch size goes to infinity and the learning rate goes to zero. The argument is as follows: as batch size goes to infinity, the estimation error goes to zero and the stochastic gradient $X^\top G$ converges\footnote{We assume the loss is computed as an average over the batch dimension, so the expression $X^\top G$ contains an implicit $1/b$ factor.} to the true gradient $\mathbb{E}[x g^\top]$. As learning rate goes to zero, the model's rate of change tends to zero, which means the rate of change of the true gradient $\mathbb{E}[x g^\top]$ tends to zero, so $M$ converges to $\mathbb{E}[x g^\top]$ and $V_{ij}$ converges to $\mathbb{E}[x g^\top]_{ij}^2$. The weight update is therefore:
\begin{align}
(W_U)_{ij} &= \frac{M_{ij}}{\sqrt{V_{ij}} + \varepsilon} \\
&= \frac{\mathbb{E}[x_i g_j]}{\sqrt{\mathbb{E}[x_i g_j]^2} + \varepsilon} \\
&\approx \frac{\mathbb{E}[x_i g_j]}{\sqrt{\mathbb{E}[x_i g_j]^2}} \\
&= \sgn(\mathbb{E}[x_i g_j])
\end{align}
This argument is simple, but I could not find it in the literature.

Although Adam performs better than sign descent in practice, I believe sign descent provides accurate intuition for how Adam works. Sign descent is simple, robust, and invariant to diagonal transformation of inputs and gradients. Adam shares all these qualities.

\subsection{Zero gradient}
SGD has the property that if $\mathbb{E}[xg^\top]=0$, then the norm of $W_U$ scales as the inverse square root of the batch size. This is desirable because the ideal update is zero in this case, and we want $W_U$ to get closer to the ideal update as the batch size increases. Iso shares this property, but Adam does not: as the batch size increases, the stochastic gradient $X^\top G$ gets smaller, but this is cancelled out by the decrease in $V$. The Adam update does not approach zero until the batch size becomes so large that the denominator of (\ref{adamupdate}) is dominated by $\varepsilon$.

More generally, this issue arises from the fact that Adam chooses a strict way of enforcing the invariant step size principle: roughly, it requires for each $i,j$ that $(W_U)_{ij}^2$ has an average value of $1$ across the past $1/(1-\beta_2)$ updates. This rule is simple and effective, but it encounters problems when the ideal value of $(W_U)_{ij}$ is small.

\subsection{Orthogonal equivariance}
Transformers (and most other architectures) have natural symmetry: if $Q$ is an orthogonal matrix, then if the input embedding is right-multiplied by $Q$, each residual branch input is left-multiplied by $Q^\top$, each residual branch output is right-multiplied by $Q$, and the output embedding is left-multiplied by $Q^\top$, we obtain a different neural network which produces the same output as the original for any input. SGD and Iso preserve this symmetry: if $A$ and $B$ are networks related in this way by some $Q$, then provided they are trained on the same data in the same order, they will produce the same output at each step of optimization. Adam lacks this symmetry because it uses elementwise scaling, which is not equivariant to orthogonal change of basis.

Although it's not immediately obvious that this lack of symmetry would cause problems, noise is generally harmful to network performance. Since $A$ and $B$ produce the same outputs for all inputs, it is reasonable to view them as the same, and therefore to consider any difference in their optimization trajectories to be noise. This suggests performance could be improved if the noise were removed.

\subsection{Linear regression}
In this subsection, I introduce a simple linear regression problem, and derive the updates made by Iso and Adam on the first step of optimization.

Let $z \sim \mathcal{N}(0, I_n)$. We define $x$, the input to our linear model, as $x=\Sigma z$, where $\Sigma\in\mathbb{R}^{n \times n}$ is a symmetric positive definite matrix. The regression target $y$ is given by $Ax$, where $A \in \mathbb{R}^{n \times n}$. Our model is $f(x) = W^\top x$, where $W \in \mathbb{R}^{n \times n}$, and the loss is $\frac{1}{2}\norm{f(x) - Ax}_2^2$. The loss is minimized when $W=A$. We are interested in the updates made by Iso and Adam at initialization, when $W=0$.

We let $Z \in \mathbb{R}^{b \times n}$ represent the values of $z$, where $b$ is the batch size, so that $X=Z\Sigma$, $Y=Z\Sigma A$, and $G = Y$ (since $W=0$, the gradient with respect to the model output is equal to the regression target). The gradient of $W$ is $X^\top G = \Sigma Z^\top Z \Sigma A$. We assume the batch size is large, so that $Z^\top Z \approx I_n$ and therefore $X^\top G = \Sigma^2 A$. Since it is the first iteration, the bias-corrected second Adam moment will be equal to the square of the first moment, so the Adam update is:
\begin{equation}
	W_U = \sgn(\Sigma^2 A)
\end{equation}

What is the Iso update in this case? Since the optimal solution is $A$, the ideal update would be $W_U=A$. However, an instance of the problem with $A = A_1$ can be converted to an instance with $A = A_2$ by applying the linear transformation $A_1^{-1}A_2$ to the gradient, so $W_U=A$ is impossible since $\norm{W_U}_F$ must be invariant to such transformations. Instead, the Iso update is:
\begin{align}
W_U &= (X^\top X)^{-1/2} X^\top G (G^\top G)^{-1/2} \\
&= (\Sigma \cancel{Z^\top Z} \Sigma)^{-1/2} \Sigma \cancel{Z^\top Z} \Sigma A (A^\top \Sigma \cancel{Z^\top Z} \Sigma A)^{-1/2} \\
&= (\Sigma \Sigma)^{-1/2} \Sigma \Sigma A (A^\top \Sigma \Sigma A)^{-1/2} \\
&= \Sigma A (A^\top \Sigma \Sigma A)^{-1/2} \\
&= B (B^\top B)^{-1/2} \text{\hspace{2em} where $B=\Sigma A$}
\end{align}
The matrix $B (B^\top B)^{-1/2}$ has a natural interpretation as the closest orthogonal matrix to $B$. This shows that in this case, Iso's first update is equal to the projection of $\Sigma A$ onto the set of orthogonal matrices.

\subsection{Summary}
Summarizing this section, the reasons to prefer Iso to Adam are the following:
\begin{enumerate}
	\item When the true gradient is zero, Iso has the correct asymptotic behavior with respect to batch size, but Adam does not.
	\item Iso is equivariant to orthogonal transformations of the weight matrices, but Adam is not.
	\item In the case of stochastic linear regression, the first Iso update is the projection of $\Sigma A$ onto the set of orthogonal matrices, while the first Adam update is $\sgn(\Sigma^2 A)$. The Iso update is a more natural mathematical operation and has linear rather than quadratic dependence on $\Sigma$.
	\item Although Iso is more complex to implement than Adam, I would argue it has shorter description length because it is a natural extension of the scalar correlation coefficient $\rho=\mathbb{E}[xy]/\sqrt{\Var(x)\Var(y)}$ to the multivariate case.
\end{enumerate}

\section{Experiments}
\label{experiments}
\subsection{Training Transformers on OpenWebText}

\begin{algorithm}
\begin{algorithmic}\itemindent=-10pt
\STATE Weight matrix $W \in \mathbb{R}^{n \times m}$
\STATE Learning rate $0 < \alpha$
\STATE Momentum decay $0 \le \beta_1 < 1$
\STATE Normalization decay $0 \le \beta_2 < 1$
\STATE $\varepsilon > 0$
\STATE Initialize first moment $M \in \mathbb{R}^{n \times m} = 0$
\STATE Initialize second moment $V \in \mathbb{R}^{n \times m} = 0$
\STATE Initialize EMA denominators $d_1,d_2 = 0$
\STATE Initialize left covariance $L \in \mathbb{R}^{n \times n} = 0$
\STATE Initialize right covariance $R \in \mathbb{R}^{m \times m} = 0$
\FOR{each iteration}
  \STATE Receive inputs $X \in \mathbb{R}^{b \times n}$ and output gradients $G \in \mathbb{R}^{b \times m}$
  \STATE $M \leftarrow M + (1 - \beta_1)(X^\top G - M)$
  \STATE $L \leftarrow L + (1 - \beta_1)(X^\top X - L)$
  \STATE $R \leftarrow R + (1 - \beta_1)(G^\top G - R)$
  \STATE $d_1 \leftarrow d_1 + (1 - \beta_1)(1 - d_1)$
  \STATE $d_2 \leftarrow d_2 + (1 - \beta_2)(1 - d_2)$
  \STATE $U := d_1 L^{-1/2} X^\top G R^{-1/2}$
  \STATE $V \leftarrow V + (1 - \beta_2)(U \odot U - V)$
  \STATE $W_{ij} \leftarrow W_{ij} - \alpha(L^{-1/2} M R^{-1/2})_{ij} / (\sqrt{V_{ij}/d_2}+\varepsilon)$
\ENDFOR
\end{algorithmic}
\caption{IsoAdam.}
\label{alg:isoadam}
\end{algorithm}

New optimizers generally require hyperparameter tuning to work well. This makes them hard to evaluate because an apparent improvement in efficiency from a new optimizer may simply be the result of its hyperparameters being tuned more carefully than the baseline \cite{kaddour2023train}.

To address this issue, I introduce IsoAdam (Algorithm \ref{alg:isoadam}), which allows hyperparameters to be transferred from Adam without re-tuning. IsoAdam multiplies the gradient on the left by $\Cov(x)^{-1/2}$ and on the right by $\Cov(g)^{-1/2}$, like Iso, then applies elementwise scaling afterward, like Adam.

\begin{figure}
\begin{center}
\includegraphics[width=0.8\textwidth]{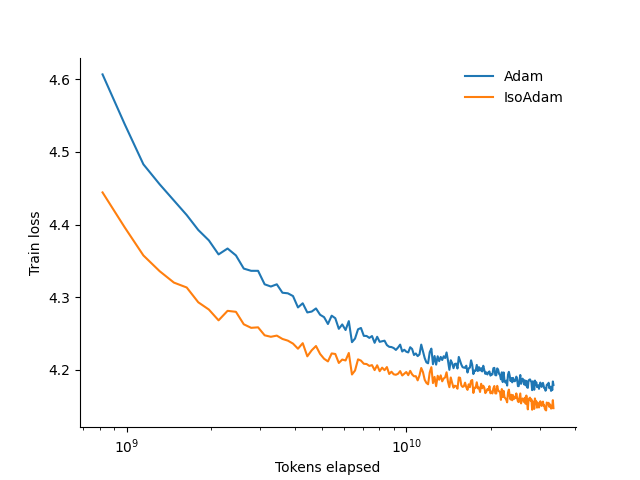}
\end{center}
\caption{Train loss on OpenWebText. Points are sampled every 1000 training iterations. Hyperparameters are identical betweeen Adam and IsoAdam, and were not modified from the standard Adam parameters.}
\label{owtloss}
\end{figure}

Using IsoAdam to train a Transformer on OpenWebText, I obtain a speedup over Adam without hyperparameter tuning, shown in Figure \ref{owtloss}. Admittedly, the model is small (around 800,000 non-embedding parameters), and more experiments are needed to validate the result at scale. IsoAdam is only used for the weight matrices; other parameters, including input and output embeddings, are optimized with Adam. Hyperparameters are given in Figure $\ref{owthps}$.

\begin{figure}
\begin{center}
\begin{tabular}{l | l }
Batch size (tokens) & 163,840 \\
\hline Width & 128 \\
\hline Number of layers & 4 \\
\hline Number of heads & 4 \\
\hline Data type & 32-bit floating point
\end{tabular}
\end{center}
\caption{Hyperparameters for IsoAdam on OpenWebText. All other hyperparameters are left at the nanoGPT default values (\url{https://github.com/karpathy/nanoGPT/blob/eba36e84649f3c6d840a93092cb779a260544d08/config/train_gpt2.py})}
\label{owthps}
\end{figure}

\subsection{Overparameterized regression}
I create a synthetic regression problem with dimension $n$ by sampling a random matrix $A\sim\mathcal{N}(0, I_{n\times n})$. The input $x$ is distributed as $x \sim \mathcal{N}(0, I_n)$ and the targets $y$ are given by $y = Ax$. The model is a sequence of $k$ linear layers: $f(x) = (W_1\dots W_k)^\top x$. Each matrix $W$ is initialized as $W \sim \frac{\mathcal{N}(0, I_{n\times n})}{\sqrt{n}}$. The model is overparameterized, since a single linear layer would be enough to learn the function.

\begin{figure}
\begin{center}
\includegraphics[width=0.8\textwidth]{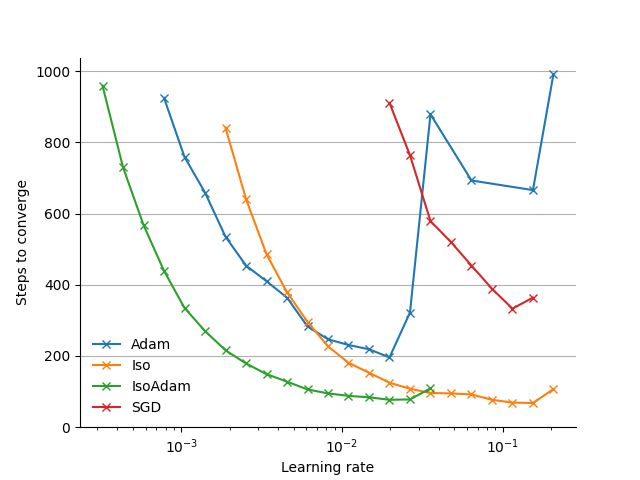}
\end{center}
\caption{Iterations to convergence with a 5-layer network. All optimizers use $\beta_1 = 0.9$. Adam and IsoAdam use $\beta_2=0.99$. Changing $\beta_2$ to 0.9 did not improve Adam's performance. ``Convergence'' means achieving mean squared error less than 1\% of a baseline that predicts all zeros. All optimizers were run with 30 learning rates logarithmically spaced between $0.5$ and $10^{-4}$. Runs that did not converge are not shown.}
\label{fig:shallow}
\end{figure}

\begin{figure}
\begin{center}
\includegraphics[width=0.8\textwidth]{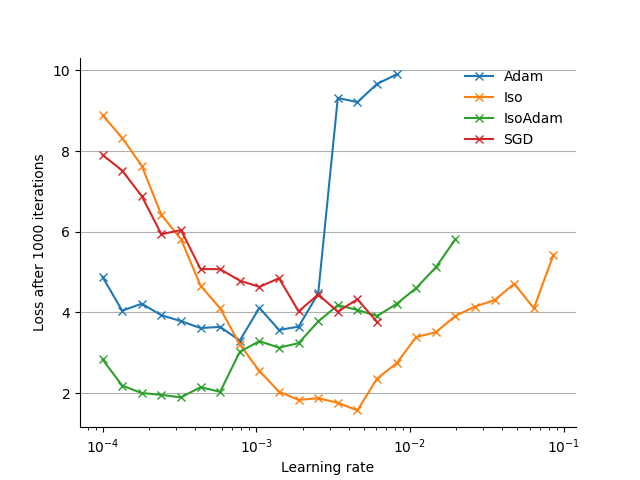}
\end{center}
\caption{Loss after training a 40-layer network for 1000 iterations. Loss is mean squared error scaled so that a baseline that predicts all zeros achieves loss 10. All hyperparameters are the same as in Figure \ref{fig:shallow}. Runs that did not achieve lower loss than the baseline are not shown.}
\label{fig:deep}
\end{figure}


To control for the learning rate, I sweep over 30 learning rates for each algorithm and plot performance with learning rate on the $x$-axis. Figure \ref{fig:shallow} plots iterations to convergence on a shallow 5-layer network. Figure \ref{fig:deep} plots loss after training for 1000 iterations with a deep 40-layer network. In both cases, Iso achieves the best performance when all optimizers use their optimal learning rate.

IsoAdam achieves performance similar to Iso while having optimal learning rate similar to Adam, validating its utility as an alternative to Adam that does not require changes to hyperparameters.

These experiments use $n=32$ and a batch size of 128.

\section{Scalability}
\label{scalability}
Can Iso scale to the largest models? I believe so. Iso requires the following computations:
\begin{enumerate}
	\item Computing the covariance matrix $X^\top X$ for the input to each layer and $G^\top G$ for the gradient of each layer's outputs.
	\item Computing the inverse square roots $L=(X^\top X)^{-1/2}$ and $R=(G^\top G)^{-1/2}$.
	\item Multiplying the gradient by $L$ and $R$: $W_U = L^\top X^\top G R$, where $X^\top G$ is already known.
\end{enumerate}
Let $b$ be the batch size in tokens and $n$ be the model width. (1) is $O(bn^2)$ while (2) and (3) are $O(n^3)$. Recent large models \cite{touvron2023llama} have batch size in tokens 100-200 times greater than width, so the cost of (1) dominates. A model with $N$ parameters using recompute in the backward pass takes about $4N$ FLOPs per token. The matrix multiplication in (1) takes around $2N$ FLOPs per token, resulting in a 50\% increase in total compute if implemented naively. However, assuming a well-behaved distribution, an $n \times n$ covariance matrix can be estimated with $O(n)$ samples, so the covariance can be computed over a subsample of $O(n)$ tokens without significant loss of accuracy. This brings the cost of (1) down to $O(n^3)$.

With sharding of all operations and an accelerator-friendly implementation of inverse matrix square root such as Newton-Schulz iteration \cite{song2022fast}, it should be possible to implement Iso with less than 10\% overhead.

\section{Related work}
\subsection{Shampoo}
The most closely related existing optimizer is Shampoo \cite{gupta2018shampoo}, which also multiplies the gradient update of a weight matrix by preconditioners on the left and right. Shampoo computes the left preconditioner as the inverse fourth root of the average of $H H^\top$, where $H$ is the gradient of the weight matrix, and the right preconditioner as the inverse fourth root of the average of $H^\top H$. Using $X$ and $G$ as defined in Section \ref{derivation}, and ignoring the fact that Shampoo computes its preconditioners using the average over the whole optimization history, we have:
\begin{align}
	\text{Shampoo:}\hspace{2.5em} &W_U = \left[ X^\top G G^\top X \right]^{-1/4} X^\top G \left[G^\top X X^\top G\right]^{-1/4} \\
	\text{Iso:}\hspace{2.5em} &W_U = \left[ X^\top X \right]^{-1/2} X^\top G \left[G^\top G\right]^{-1/2}
\end{align}
If $X$ and $G$ are symmetric, positive definite, and simultaneously diagonalizable, these expressions both equal $I_n$.

Shampoo has a natural generalization to tensors of any rank. Iso also has such a generalization. If $W\in\mathbb{R}^{n\times m \times p}$ is a tensor with inputs $a\in\mathbb{R}^n$ and $b\in\mathbb{R}^m$ that computes its output as $c_k=\sum_{i=1}^n \sum_{j=1}^m W_{ijk} a_i b_j$, and the gradient with respect to $c$ is $g$, then the Iso preconditioner multiplies by $\Cov(a)^{-1/2}$ along the first dimension, $\Cov(b)^{-1/2}$ along the second dimension, and $\Cov(g)^{-1/2}$ along the third dimension.

\subsection{Analysis and extensions of Adam}
Many explanations have been proposed for Adam's success. One explanation is that it estimates the Hessian \cite{molybog2023theory}. Another is that it is robust to heavy-tailed noise \cite{zhang2020adaptive}. Many papers \cite{bernstein2018signsgd,kunstner2023noise,balles2020dissecting} have connected Adam to sign descent, also known as SignSGD \cite{bernstein2018signsgd}.

Various modifications of Adam have been proposed to achieve better performance \cite{chen2023symbolic,zhuang2020adabelief} or memory efficiency \cite{shazeer2018adafactor}.

\subsection{Second-order optimizers}
Many optimizers seek to use the Hessian to take shorter steps along high-curvature directions. Because the Hessian has size quadratic in the number of network parameters, it is necessary to approximate it, for example using diagonal \cite{liu2023sophia} or block-factored \cite{martens2020optimizing} approximatons.

\subsection{AdaGrad and approximations}
Full-matrix AdaGrad \cite{JMLR:v12:duchi11a} multiplies the gradient by the inverse square root of its covariance matrix. Approximating the covariance matrix as diagonal yields an algorithm similar to Adam. GGT \cite{agarwal2020efficient} uses a non-diagonal low-rank approximation of the covariance matrix.

\bibliography{iso}

\end{document}